\documentclass{article}

\pdfoutput=1

\usepackage{graphicx}

\usepackage[final]{neurips_2024}

\usepackage[utf8]{inputenc} 
\usepackage[T1]{fontenc}    
\usepackage{hyperref}       
\usepackage{url}            
\usepackage{booktabs}       
\usepackage{amsfonts}       
\usepackage{nicefrac}       
\usepackage{microtype}      
\usepackage{xcolor}         
\usepackage{svg}  
\usepackage{amsmath, amssymb, amsthm}

\newtheorem{defn}{Definition}

\title{
Peter Parker or Spiderman?\\
Disambiguating Multiple Class Labels 
}

\author{%
  Nuthan Mummani$^1$, Simran Ketha$^{1,2}$, Venkatakrishnan Ramaswamy$^{1,2}$  \thanks{ Code available in \href{https://github.com/mummani-nuthan/Disambiguating-Multiple-Class-Labels}{https://github.com/mummani-nuthan/Disambiguating-Multiple-Class-Labels}} \\
  $^1$Department of Computer Science \& Information Systems, 
  \\Birla Institute of Technology \& Science Pilani, Hyderabad 500078, India. \\
  $^2$Anuradha \& Prashanth Palakurthi Centre for Artificial Intelligence Research, \\
  Birla Institute of Technology \& Science Pilani, Hyderabad 500078, India. \\
 \texttt{\{h20221030057, p20200021, venkat\}@hyderabad.bits-pilani.ac.in} \\
}

\begin{document}

\maketitle

\begin{abstract}

  In the supervised classification setting, during inference, deep networks typically make multiple predictions.
For a pair of such predictions (that are in the top-$k$ predictions), two distinct possibilities might occur. On the one hand, each of the two predictions might be primarily driven by two distinct sets of entities in the input. On the other hand, it is possible that there is a single entity or set of entities that is driving the prediction for both the classes in question. This latter case, in effect, corresponds to the network making two separate guesses about the identity of a single entity type. Clearly, both the guesses cannot be true, i.e. both the labels cannot be present in the input.
Current techniques in interpretability research do not readily disambiguate these two cases, since they typically consider input attributions for one class label at a time. 
 Here, we present a framework and method to do so, leveraging modern segmentation and input attribution techniques. Notably, our framework also provides a simple counterfactual “proof” of each case, which can be verified for the input on the model (i.e. without running the method again). We demonstrate that the method performs well for a number of samples from the ImageNet validation set and on multiple models.
  
\end{abstract}

\section{Introduction}

Supervised deep learning models performing classification are being widely deployed in many settings. An important and active direction of research is on interpretability of the predictions of these models. In the multiclass classification setting, typically, each training datapoint comes with one or few labels \cite{5206848}, \cite{lin2014microsoft}; however models usually output softmax prediction "probabilities" for every class label present in the dataset. Generally, either the top one or top $k$ softmax values are considered as predictions for classification. Since contemporary datasets have large numbers of labels, many of the labels in the top $k$ predictions are likely those that aren't present in the input in question. Indeed, for a given pair of such predicted labels, these prediction probabilities have two distinct interpretations. The first interpretation is that the probabilities represent the possibility of the presence of distinct entities that correspond to each of the class labels. The second interpretation is that the pair of probabilities represent two distinct predictions about a single type of entity present in the image. Both these interpretations could simultaneously be true for different pairs of predicted class labels for a single input that is run through a model. The second interpretation being true for a given pair of labels might detract from our confidence that both the labels are indeed correct predictions; this would indicate the need to verify these predictions via other means -- e.g. using a different more capable model or a human. Most contemporary models do not offer a direct way to disambiguate these two interpretations for a given pair of top $k$ prediction probabilities, for a single input datapoint. To our knowledge, no existing techniques in the literature have been explicitly designed to address this issue.

In this paper, we build a framework and method to address this problem. On one hand, we build a counterfactual proof framework that entails setting up definitions that disambiguate the aforementioned types of class label pairs. Specifically, we stipulate that an assertion about whether a given pair of labels correspond to distinct entity-types or to a single type of entity must be accompanied by a counterfactual proof, which is a certificate that can be used to verify this claim using the model. Notably, verifying the claim does not require re-running the method that produced it, or indeed even knowledge of the method. Conceptually, this is a departure from typical attribution methods, wherein there isn't an explicit way to objectively verify a claimed attribution that is divorced from the technique that produced it. Secondly, for image recognition, using modern segmentation and attribution techniques, we propose methods that produces such a counterfactual proof. To this end, we first segment the given input image and using existing attribution techniques assign segment-wise attribution scores for each label. These scores are used to determine if the two given label predictions point to the same set of entities or different set of entities. We then build counterfactual proofs that satisfy the aforementioned definitions. Using a number of images from the ImageNet validation set, we demonstrate that the method performs favorably.

\section{Related work}
\label{relatedwork}

Attribution techniques have been studied in multiple directions. Perturbations are the simplest among the them. \cite{zeiler2014visualizing} implements them by masking the part of the picture with gray square and observing the output. They also implemented a process where outputs at each layer are projected back to the layer’s input dimension with minimum loss in the data but also capturing what caused the final activation and termed it as deconvolution. \cite{springenberg2014striving} proposed Guided backpropagation, as a modification to deconvolution.

Gradient-based methods, such as gradient descent and backpropagation, form the foundation for many feature attribution techniques. These methods compute the gradient of the model's output with respect to the input features. The magnitude of the gradient indicates the sensitivity of the model's output towards changes in the input features. Gradient \cite{simonyan2014deepinsideconvolutionalnetworks} itself along with integrated gradients \cite{sundararajan2017axiomatic}, deepLift \cite{pmlr-v70-shrikumar17a}, GradCAM \cite{selvaraju2017grad}, layer wise relevance propagation \cite{bach2015pixel} are few notable techniques.

 By closely inspecting the visualizations of these gradients, \cite{sundararajan2017axiomatic} proved that gradients do not work properly. XRAI \cite{kapishnikov2019xrai} showcases the ability to attribute to particular segments of image with the help of Integrated gradients and segmentation algorithms like Felzenswalb’s graph based algorithm \cite{felzenszwalb2004efficient}.

LIME \cite{ribeiro2016should} and SHAP \cite{lundberg2017unified} have different approaches. LIME tried to generate interpretable explanations local to the input in question which are understandable to humans. SHAP on the other hand tries to explains how important the feature is in the given prediction based on a concept of Shapley values from Game Theory. While all these methods calculate/explain the importance of each feature for the given generated output, we would like to expand on the role of these features when we consider multiple outputs.

Relatedly, the issue of popular contemporary datasets such as ImageNet having one label per image has also received attention. For example, \cite{beyer2020imagenet} point out that even though images in ImageNet training set often contain multiple objects, only one of them is recognized in the label. 

\section{Definitions and Preliminaries}
\label{defs}
We now present some definitions and preliminaries that will be used in the remainder of the paper. While we apply the framework to the image classification setting here, these definitions could, in principle, also be applied to other types of supervised learning models. 

For our purposes here, we define a deep network model as a function that maps input points in $n$-dimensional space to a vector of softmax “probabilities” corresponding to $m$ class labels.

\begin{defn}
A deep network model is a function $f:\mathbb{R}^n\rightarrow[0,1]^m$ , which maps an input in $n$-dimensional space to a vector of softmax values corresponding to $m$ class labels.
\end{defn}

Next, we define the notion of a {\em redaction} of an input, which intuitively corresponds to replacing values in a subset $S$ of the dimensions in the input to a single fixed value. The idea is that doing so will remove the information present in those dimensions and such a redacted input would serve as a counterfactual input. While we do not claim that this manner of constructing counterfactual inputs is a canonical one, we find that it does work well in practice for image recognition networks, as demonstrated in Section \ref{effectiveredactions}.
\begin{defn}[$S$-redaction]
Given an input $I\in\mathbb{R}^n$ and a set $S\subseteq \{1, \ldots, n\}$ of indices, an {$S$-redaction} of $I$ to $v$, is defined as the input $I_S$ obtained by replacing the values corresponding to the indices in $S$ to the value $v$. 
\end{defn}
Unless otherwise specified, when we mention an $S$-redaction here, we mean an $S$-redaction to zero. If the input is an image, an $S$-redaction of it would correspond to the image generated by "blackening" out the subset of the pixels corresponding to S. Also, if each pixel has multiple channels (e.g. R,G,B), an $S$-redaction will zero out values in all channels, for every pixel present in S.

Informally, an input attribution for a specific class label is typically understood to correspond to the input dimensions that are "responsible" for the prediction of that class label by the deep network model. Here, we will define an attribution to simply be a subset of input dimensions (i.e. without assigning relative weights to every dimension in the subset). We now define a natural counterfactual notion of input attributions that precisely quantifies the same in a verifiable manner.

\begin{defn}[$\delta$-attribution]
\label{def4}
For a deep network  $f:\mathbb{R}^n\rightarrow[0,1]^m$, input  $I\in\mathbb{R}^n$, label $l$ with prediction $p$, $\delta \in[0,1]$, and $S \subseteq \{1, …, n\}$, if the S-redaction of $I$ causes the prediction of ~$l$ to be at most $\delta p$, then $S$ is said to be a $\delta$-attribution for label $l$ corresponding to input $I$, with respect to f.
\end{defn}

Here, the intent is to have $\delta$ be a small value (e.g. $\delta=0.2$)

This definition of a $\delta$-attribution naturally leads to a verification method. The idea is that one can accompany a claimed $\delta$-attribution with a counterfactual proof or certificate, which in this case would simply be the $\delta$-attribution $S$. This allows a verifier to easily verify a claimed $\delta$-attribution without needing to re-run the method that determined it or indeed even having knowledge of the method.

We now define the two major types of label predictions, given a pair of label predictions for an input by a deep network, namely $\delta$-disjoint label predictions and $\delta$-overlapping label predictions. A $\delta$-disjoint label prediction corresponds to the case in which two distinct types of entities are present in the input that correspond respectively to the two class labels in question, with softmax prediction values being $p_1$ and $p_2$ respectively. The definition posits that if this is the case, then there must exist two redactions -- an $S_1$-redaction and an $S_2$-redaction -- where $S_1$ and $S_2$ are disjoint sets. Furthermore, the $S_1$-redaction must cause the softmax prediction for the first label to dip to $\delta p_1$ or below, while keeping the softmax prediction for the second label to be at least $(1-\delta)p_2$. The $S_2$-redaction behaves likewise for the second label.
\begin{defn}[$\delta$-disjoint label predictions]
\label{type1def}
For $\delta \in [0,0.5]$, suppose we have a deep network  $f:\mathbb{R}^n\rightarrow[0,1]^m$ which, on input $I$, has predictions $p_1$ and $p_2$ for class labels $l_1$ and $l_2$ respectively. The class labels $l_1$ and $l_2$ are said to be {\em $\delta$-disjoint}, if there exist disjoint sets $S_1$ and $S_2$ such that  
\begin{enumerate}
\item The $S_1$-redaction of $I$ causes a $\delta$-attribution to exist for class label $l_1$, while causing the prediction for class $l_2$ to be at least $(1-\delta)p_2$. 
\item The $S_2$-redaction of $I$ causes a $\delta$-attribution to exist for class label $l_2$, while causing the prediction for class $l_1$ to be at least $(1-\delta)p_1$. 
\end{enumerate}
\end{defn}

Here, again, for two labels $l_1$ and $l_2$, claimed $\delta$-disjoint label predictions will be accompanied by a certificate, which would simply be the $\delta$-attributions $S_1$ and $S_2$ that satisfy the above definition.

Next, given a pair of label predictions for an input by a deep network, we define $\delta$-overlapping label predictions. This is the case when the two labels in question correspond to a single entity type present in the input.  The idea is to establish this case by demonstrating a $S$-redaction that is a $\delta$-attribution for both the class labels without the labels being $\delta$-disjoint.

\begin{defn}[$\delta$-overlapping label predictions]
\label{type2def}
For a deep network  $f:\mathbb{R}^n\rightarrow[0,1]^m$ with an input $I$, two class labels $l_1$ and $l_2$ are said to be {\em $\delta$-overlapping}, if  $l_1$ and $l_2$ are not $\delta$-disjoint and if there exists a set $S$ such that an $S$-redaction causes a $\delta$-attribution to exist for class labels $l_1$ as well as $l_2$. 
\end{defn}

Here, again, the certificate would be the $\delta$-attribution $S$; however it is unclear if a tractable verification algorithm exists, since one might need to check all partitions of S -- of which there are exponentially many -- to check if they correspond to $\delta$-disjoint label predictions. In Section \ref{type1algo3}, we describe a heuristic verification algorithm that is tractable and demonstrate that it works well.


\begin{figure}[h!]
    \centering
        \includegraphics[width=0.99\linewidth]{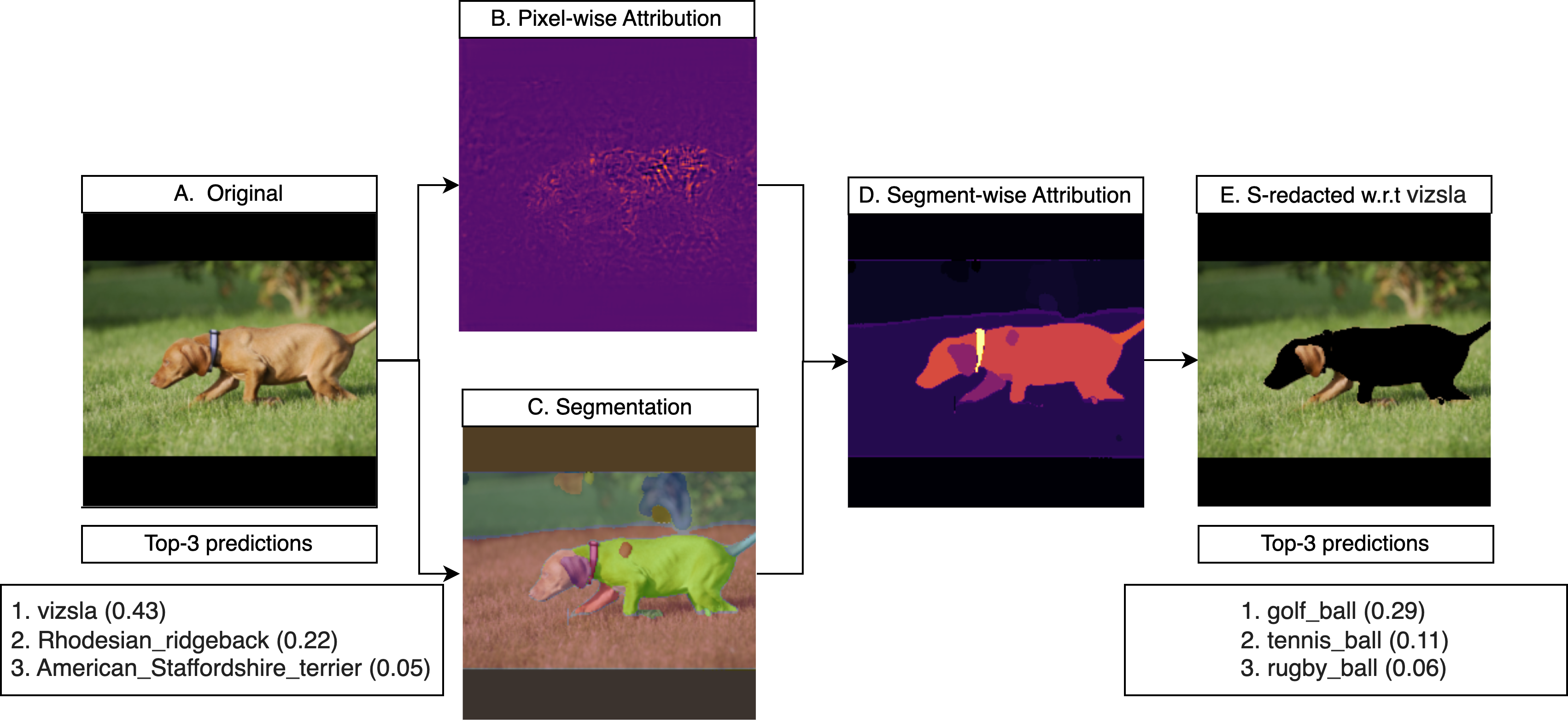}
        \caption{An illustration of rank-based redaction. {\bf A.} An image from the ImageNet validation set from the class {\tt vizsla} is padded with zeros to match the input dimension of VGG16 model to obtain the image shown. Corresponding top-3 predictions are mentioned. {\bf B.} The image in A. is attributed to the label {\tt vizsla} using integrated gradients to obtain pixelwise attribution values. {\bf C.} The image in A. is segmented using the SAM model. {\bf D.} The pixel-wise attribution values from B. are averaged over the segments and these segments are ranked accordingly to get segment-wise attributions for the label {\tt vizsla}. {\bf E.} Top 25\% of the ranked segments are then redacted to get an $S$-redaction. Corresponding top-3 predictions for this $S$-redacted image are mentioned. The prediction for {\tt vizsla} on this $S$-redacted image dropped to 0.010. This process on the same image with ResNet-50 and Inception-v3 are shown in Figure~\ref{overview_resnet_inception}.}
    \label{fig1}
\end{figure}

\section{Methodology}
\label{Redaction}
Leveraging modern input attribution and segmentation techniques, we build algorithms to determine if a given pair of labels is $\delta$-disjoint or $\delta$-overlapping. These algorithms also return the corresponding certificates. We deploy and test these algorithms on image classification models VGG-16 \cite{simonyan2014very}, Inception-v3 \cite{szegedy2016rethinking}, and ResNet-50 \cite{he2016deep} which are pretrained on the ImageNet dataset \cite{5206848}. We use images from the ImageNet validation dataset in our test, unless otherwise mentioned.

For a label available in the top $k$ predictions of an input image, we calculate pixel-wise attribution using integrated gradients \cite{sundararajan2017axiomatic} and parallelly, we segment the image using Segment Anything Model (SAM) \cite{kirillov2023segment}. We then performed segment-wise accumulation of attribution values to rank the segments from highest attribution to lowest attribution, along the lines of XRAI \cite{kapishnikov2019xrai}. These segment-wise rankings are used in the later part of the paper and can be visualized using heatmaps (Figure \ref{fig1}).

\subsection{Effectiveness of Redactions}
\label{effectiveredactions}
Here, we demonstrate that redactions to zero are an effective counterfactual proof, in practice.
Redacted images are constructed by picking up segments one-by-one based on segment-wise attribution rankings and replacing segmented areas with black pixels in the original preprocessed image. The process of creating the redacted images for the top predicted label using VGG-16 model is shown in Figure~\ref{fig1}, for a sample image.


Every network has its own pre-processing stage where e.g. in VGG16, it zero-centers the data with respect to the dataset. Although, the redacted images shown in this paper are generated by performing redactions on images before pre-processing step for the purpose of visualization, in practice we redact the segment after the pre-processing step. 

\begin{figure}[h!]
    \centering
        \includegraphics[scale=0.68]{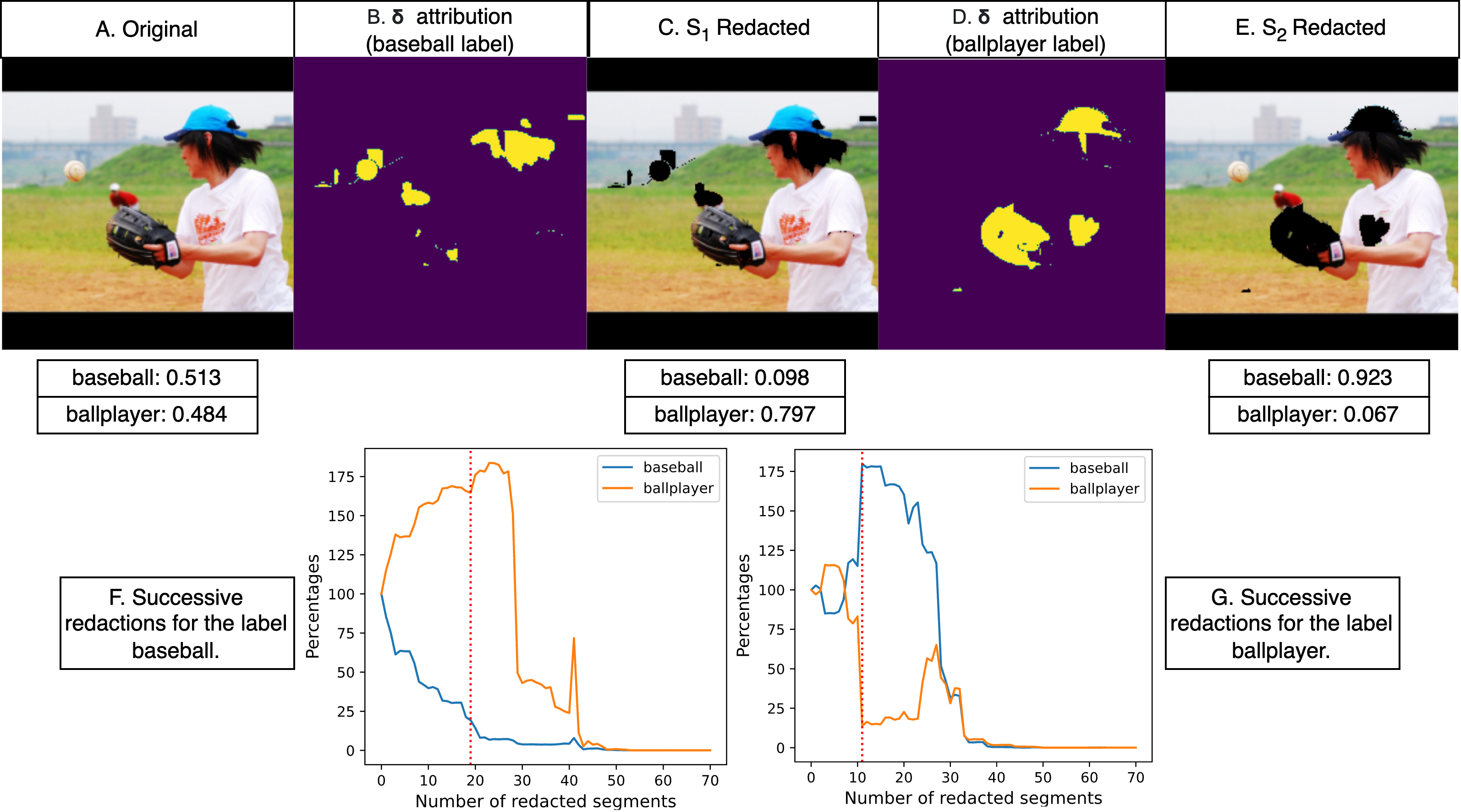}
        \caption{Example illustrating $\delta$-disjoint attributions. {\bf A.} An image from the ImageNet validation set \& its corresponding top-2 labels with their predictions on VGG-16 model. {\bf B.} For $\delta=0.2$, $\delta$-attribution for the label {\tt baseball} (indicated in {\em yellow}) obtained using the algorithm discussed in Section \ref{type1}. {\bf C.} The corresponding redacted image for the label {\tt baseball} with the resultant prediction values. {\bf D.} For $\delta=0.2$, $\delta$-attribution for the label {\tt ballplayer} (indicated in {\em yellow}) obtained using the algorithm discussed in Section \ref{type1}. {\bf E.} The corresponding redacted image for the label {\tt ballplayer} with the resultant prediction values. 
        For both labels, percentage of softmax prediction values while redacting segments with respect to the original image are plotted. {\bf F.} For the algorithm discussed in Section \ref{type1}, we plotted the percentage change in prediction for the two labels, when the segments ranked for the label {\tt baseball} were successively redacted in order of their rank. 
        Here, the $\delta=0.2$ attribution is obtained at the 19th redaction ({\em red-dotted line}) where prediction of {\tt baseball} is atmost $\delta$\*$p_1$ ($0.098 < 0.2*0.513$) and prediction of {\tt ballplayer} is atleast (1-{\em$\delta$})$p_2$ ($0.797 > 0.8*0.484$). {\bf G.} Corresponding plot for {\tt ballplayer}. Additional examples are provided in Appendix \ref{additional_examples}.}
        
    \label{fig2}
\end{figure}

\section{Distinct labels pointing to distinct entities}
\label{type1}
To determine if two labels from the top-$k$ predictions are driven by distinct entities in an image, we need $S_1$ and $S_2$ redactions, if available, that satisfy Definition \ref{type1def}. One method to obtain such redactions is discussed below \& two other methods are presented in the Appendix \ref{othertype1}. 

Given a list of segment attribution values for one label, for each segment, we determine the proportion of the segment's attribution value with respect to the highest segment attribution value, which we call its {\em normalized segment attribution}. 
We do so for the other label as well.
Now we segregate the segments into two disjoint sets corresponding to the two labels. For any segment, if the normalized segment attribution for label $l_1$ is higher than that for label $l_2$, then that segment is categorized within the set of label $l_1$ and vice-versa. In case of a tie, we use the data of surrounding segments for categorization. Now that we have two disjoint sets of segments, one for each label, we pick each segment from label $l_1$'s set based on their rank and redact by sequentially accumulating them to form an $S_1$-redaction until the prediction of corresponding class label $l_1$ goes down to at most {\em $\delta$}$p_1$ while the $l_2$ prediction stays above (1-{\em $\delta$})$p_2$ where $p_1$ and $p_2$ are the softmax probability of original image of labels $l_1$ and $l_2$ respectively. This step is repeated on $l_2$'s set to obtain an $S_2$ redaction. Redacting based on their rank allows us to get the $S_1$  and $S_2$ redactions that have a small number of segments. We find that this method is effective in finding redactions 
that satisfy Definition \ref{type1def}. In Figure \ref{fig2}, we illustrate this method, for a sample image with $\delta = 0.2$.

We explored two more ways to generate redactions that satisfy Definition \ref{type1def}. One includes finding $S$-redactions and then making them disjoint and the other generates $S$-redactions without using the pixel-wise or segment-wise attributions. These are  discussed in Appendix\ref{othertype1}. 

\section{Distinct labels pointing to single entity }
\label{type2}

\begin{figure}[!h]
    \centering
        \includegraphics[scale=0.6]{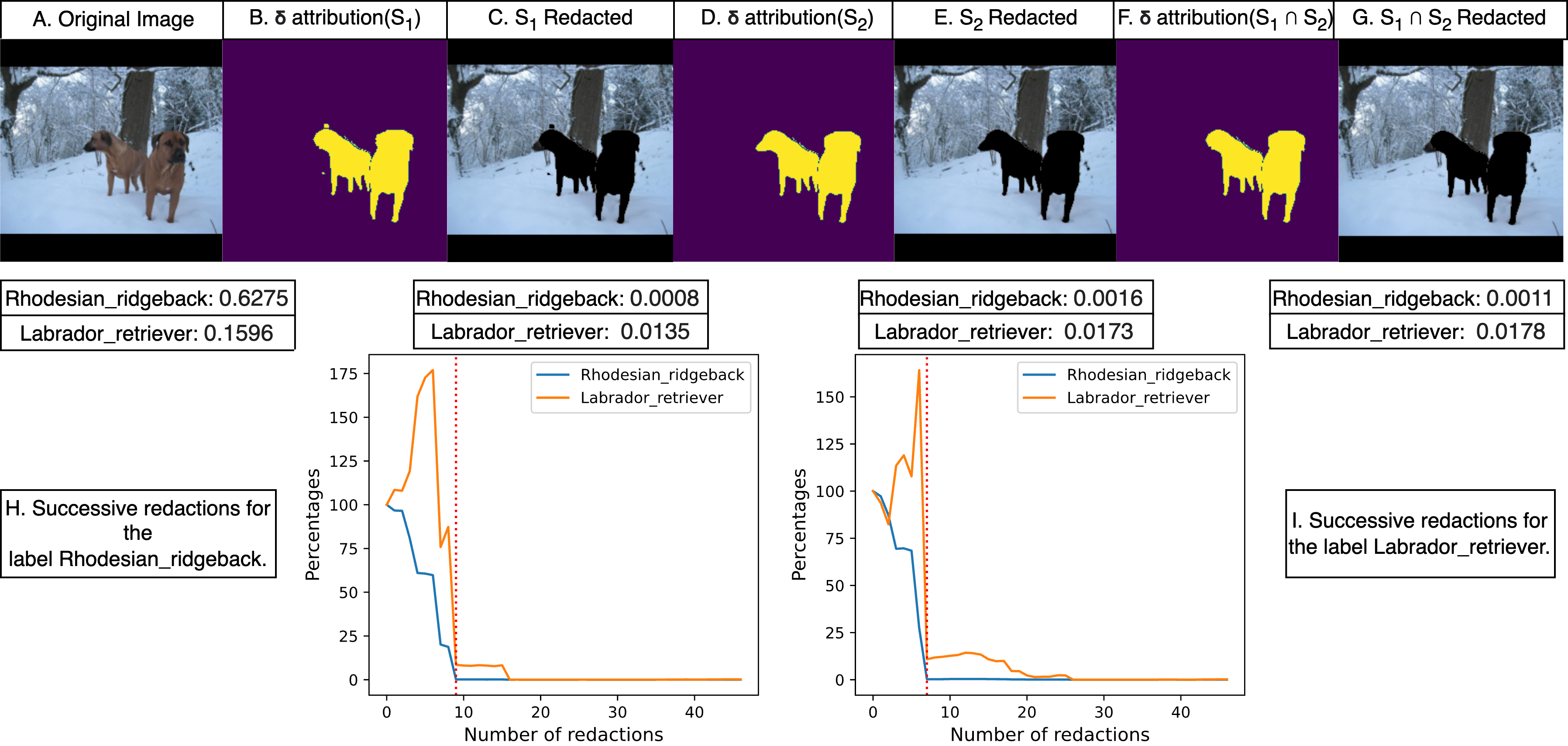}
        \caption{Example illustrating $\delta$-overlapping attributions. {\bf A.} An image from the ImageNet validation set (whose correct label from the validation set is {\tt Rhodesian\_ridgeback}) and its corresponding top-2 labels with their predictions values on VGG-16.
        {\bf B.} For $\delta=0.2$, $\delta$-attribution for the label {\tt Rhodesian\_ridgeback} (indicated in {\em yellow}) obtained using the algorithm discussed in Section \ref{type2}. {\bf C.} The corresponding redacted image for the label {\tt Rhodesian\_ridgeback} with the resultant prediction values. {\bf D.} For $\delta=0.2$, $\delta$-attribution for the label {\tt Labrador\_retriever} (indicated in {\em yellow}) obtained using the algorithm discussed in Section \ref{type2}. {\bf E.} The corresponding redacted image for the label {\tt ballplayer} with the resultant prediction values. {\bf F. } For $\delta=0.2$ , $S_1 \cap S_2$ is verified to satisfy Definition \ref{type2def}.  {\bf G. } The corresponding $S_1 \cap S_2$-redacted image with the resultant prediction values.
        For both labels, percentage of softmax prediction values while redacting segments with respect to the original image are plotted. 
        {\bf H.} For the algorithm discussed in Section \ref{type2}, we plotted the percentage change in prediction for the two labels, when the segments ranked for the label {\tt Rhodesian\_ridgeback} were successively redacted in order of their rank.
        Here $\delta=0.2$ attribution is obtained at the 9th redaction ({\em red dotted line}) where prediction of {\tt Rhodesian\_ridgeback} is atmost $\delta$\*$p_1$ ($0.0008 < 0.2*0.6275$) and prediction of {\tt Labrador\_retriever} is also atmost $\delta$\*$p_2$ ($0.0135 < 0.2*0.1596$). {\bf I.} Corresponding plot for the label {\tt Labrador\_retriever}. Additional examples are provided in Appendix \ref{additional_examples}.}

    \label{fig3}
\end{figure}

To determine if two labels from the top-$k$ predicted labels are "pointing" to a single entity in an image, we do the following. We pick each segment based on their absolute\footnote{i.e. non-normalized, as described in Section \ref{Redaction}} rank for label $l_1$ and redact by sequentially accumulating them to form an $S_1$ redaction until the prediction of both the labels go down simultaneously to at most $\delta p_1$ and  $\delta p_2$ respectively. This process is likewise repeated with segments ranked based on $l_2$ to obtain a $S_2$ redaction. If $l_1$ and $l_2$ are indeed pointing to single entity, then $S_1$ and $S_2$ redactions present themselves with a significant intersection and $S_1 \cap S_2$, on satisfying Definition \ref{type2def}, is used as a $\delta$-attribution.The $\delta$-attribution for $S_1 \cap S_2$, as illustrated in Figure~\ref{fig3}, acts as a certificate which can be used to verify that the two labels indeed "point" to a single entity in the image.

This leaves open the possibility that one chooses a pair of labels that are in fact $\delta$-disjoint and by merely generating $S=S_1 \cup S_2$ 
provides a purported certificate for $\delta$-overlapping label predictions. In such cases, each subset of the $S$ provided needs to be redacted to check 
if there exists any subset of $S$ that only causes one label to rise rather than both. But, such a check would take exponential time with respect to the number of segments, which will be intractable in most cases. Hence, to avoid this, we propose a tractable heuristic verification algorithm in Appendix \ref{type1algo3} that does not require attribution values. It separates $S_1$ and $S_2$ from $S=S_1 \cup S_2$ when initialised with $S$ and invalidates such a purported certificate. We demonstrate that this heuristic works well in Figure \ref{verification}.
To avoid generating such an incorrect $S$-redaction ourself, we first run the $\delta$-disjoint label predictions algorithm \& then run the $\delta$-overlapping predictions algorithm, if the former algorithm doesn't succeed.

\section{Discussion}
\label{Discussion}
In this paper, we consider the problem of disambiguating the input attributions of a given pair of class labels. Specifically, we ask if the two label predictions arise from the same percept or from different percepts present in the input. We build a method and framework to do so, by leveraging modern attribution and segmentation techniques and demonstrate favorable performance on a number of contemporary image classification models. 

This work comes with some limitations. Firstly, we use existing attribution and segmentation algorithms and, as such, depend on their performance; this also has the positive effect that improvements in such techniques will likely improve our method. Another limitation is that, for cases wherein the object corresponding to a label isn't present, our method does not specifically identify that this is so; see Section \ref{additional_examples}. Finally, we find, empirically, that for labels whose softmax values are very small, the method often does not perform well. Indeed, this may be because for such small prediction values the model does not tangibly use a coherent set of segments for such predictions.

A conceptual departure from typical attribution methods is our stipulation that a claimed answer ought to be accompanied by a certificate that can be objectively verified, i.e. without appeal to the method that created it. Often, different attribution methods offer differing attributions and it is difficult to objectively and automatically ascertain the quality of these attributions without human scoring. We therefore suggest that this type of framework will also have value in such settings.


\begin{ack}
Simran Ketha was supported by an APPCAIR Fellowship, from the Anuradha \& Prashanth Palakurthi Centre for Artificial Intelligence Research. The authors declare no competing interests.

\end{ack}

\medskip
{
\small
\bibliographystyle{ACM-Reference-Format}
\bibliography{references}


\begin{thebibliography}{18}


\ifx \showCODEN    \undefined \def \showCODEN     #1{\unskip}     \fi
\ifx \showDOI      \undefined \def \showDOI       #1{#1}\fi
\ifx \showISBNx    \undefined \def \showISBNx     #1{\unskip}     \fi
\ifx \showISBNxiii \undefined \def \showISBNxiii  #1{\unskip}     \fi
\ifx \showISSN     \undefined \def \showISSN      #1{\unskip}     \fi
\ifx \showLCCN     \undefined \def \showLCCN      #1{\unskip}     \fi
\ifx \shownote     \undefined \def \shownote      #1{#1}          \fi
\ifx \showarticletitle \undefined \def \showarticletitle #1{#1}   \fi
\ifx \showURL      \undefined \def \showURL       {\relax}        \fi
\providecommand\bibfield[2]{#2}
\providecommand\bibinfo[2]{#2}
\providecommand\natexlab[1]{#1}
\providecommand\showeprint[2][]{arXiv:#2}

\bibitem[Bach et~al\mbox{.}(2015)]%
        {bach2015pixel}
\bibfield{author}{\bibinfo{person}{Sebastian Bach}, \bibinfo{person}{Alexander Binder}, \bibinfo{person}{Gr{\'e}goire Montavon}, \bibinfo{person}{Frederick Klauschen}, \bibinfo{person}{Klaus-Robert M{\"u}ller}, {and} \bibinfo{person}{Wojciech Samek}.} \bibinfo{year}{2015}\natexlab{}.
\newblock \showarticletitle{On pixel-wise explanations for non-linear classifier decisions by layer-wise relevance propagation}.
\newblock \bibinfo{journal}{\emph{PloS one}} \bibinfo{volume}{10}, \bibinfo{number}{7} (\bibinfo{year}{2015}), \bibinfo{pages}{e0130140}.
\newblock


\bibitem[Beyer et~al\mbox{.}(2020)]%
        {beyer2020imagenet}
\bibfield{author}{\bibinfo{person}{Lucas Beyer}, \bibinfo{person}{Olivier~J. Hénaff}, \bibinfo{person}{Alexander Kolesnikov}, \bibinfo{person}{Xiaohua Zhai}, {and} \bibinfo{person}{Aäron van~den Oord}.} \bibinfo{year}{2020}\natexlab{}.
\newblock \bibinfo{title}{Are we done with ImageNet?}
\newblock
\newblock
\showeprint[arxiv]{2006.07159}~[cs.CV]
\urldef\tempurl%
\url{https://arxiv.org/abs/2006.07159}
\showURL{%
\tempurl}


\bibitem[Deng et~al\mbox{.}(2009)]%
        {5206848}
\bibfield{author}{\bibinfo{person}{Jia Deng}, \bibinfo{person}{Wei Dong}, \bibinfo{person}{Richard Socher}, \bibinfo{person}{Li-Jia Li}, \bibinfo{person}{Kai Li}, {and} \bibinfo{person}{Li Fei-Fei}.} \bibinfo{year}{2009}\natexlab{}.
\newblock \showarticletitle{ImageNet: A large-scale hierarchical image database}. In \bibinfo{booktitle}{\emph{2009 IEEE Conference on Computer Vision and Pattern Recognition}}. \bibinfo{pages}{248--255}.
\newblock


\bibitem[Felzenszwalb and Huttenlocher(2004)]%
        {felzenszwalb2004efficient}
\bibfield{author}{\bibinfo{person}{Pedro~F Felzenszwalb} {and} \bibinfo{person}{Daniel~P Huttenlocher}.} \bibinfo{year}{2004}\natexlab{}.
\newblock \showarticletitle{Efficient graph-based image segmentation}.
\newblock \bibinfo{journal}{\emph{International journal of computer vision}}  \bibinfo{volume}{59} (\bibinfo{year}{2004}), \bibinfo{pages}{167--181}.
\newblock


\bibitem[He et~al\mbox{.}(2016)]%
        {he2016deep}
\bibfield{author}{\bibinfo{person}{Kaiming He}, \bibinfo{person}{Xiangyu Zhang}, \bibinfo{person}{Shaoqing Ren}, {and} \bibinfo{person}{Jian Sun}.} \bibinfo{year}{2016}\natexlab{}.
\newblock \showarticletitle{Deep residual learning for image recognition}. In \bibinfo{booktitle}{\emph{Proceedings of the IEEE conference on computer vision and pattern recognition}}. \bibinfo{pages}{770--778}.
\newblock


\bibitem[Kapishnikov et~al\mbox{.}(2019)]%
        {kapishnikov2019xrai}
\bibfield{author}{\bibinfo{person}{Andrei Kapishnikov}, \bibinfo{person}{Tolga Bolukbasi}, \bibinfo{person}{Fernanda Vi{\'e}gas}, {and} \bibinfo{person}{Michael Terry}.} \bibinfo{year}{2019}\natexlab{}.
\newblock \showarticletitle{Xrai: Better attributions through regions}. In \bibinfo{booktitle}{\emph{Proceedings of the IEEE/CVF international conference on computer vision}}. \bibinfo{pages}{4948--4957}.
\newblock


\bibitem[Kirillov et~al\mbox{.}(2023)]%
        {kirillov2023segment}
\bibfield{author}{\bibinfo{person}{Alexander Kirillov}, \bibinfo{person}{Eric Mintun}, \bibinfo{person}{Nikhila Ravi}, \bibinfo{person}{Hanzi Mao}, \bibinfo{person}{Chloe Rolland}, \bibinfo{person}{Laura Gustafson}, \bibinfo{person}{Tete Xiao}, \bibinfo{person}{Spencer Whitehead}, \bibinfo{person}{Alexander~C Berg}, \bibinfo{person}{Wan-Yen Lo}, {et~al\mbox{.}}} \bibinfo{year}{2023}\natexlab{}.
\newblock \showarticletitle{Segment anything}. In \bibinfo{booktitle}{\emph{Proceedings of the IEEE/CVF International Conference on Computer Vision}}. \bibinfo{pages}{4015--4026}.
\newblock


\bibitem[Lin et~al\mbox{.}(2014)]%
        {lin2014microsoft}
\bibfield{author}{\bibinfo{person}{Tsung-Yi Lin}, \bibinfo{person}{Michael Maire}, \bibinfo{person}{Serge Belongie}, \bibinfo{person}{James Hays}, \bibinfo{person}{Pietro Perona}, \bibinfo{person}{Deva Ramanan}, \bibinfo{person}{Piotr Doll{\'a}r}, {and} \bibinfo{person}{C~Lawrence Zitnick}.} \bibinfo{year}{2014}\natexlab{}.
\newblock \showarticletitle{Microsoft coco: Common objects in context}. In \bibinfo{booktitle}{\emph{Computer Vision--ECCV 2014: 13th European Conference, Zurich, Switzerland, September 6-12, 2014, Proceedings, Part V 13}}. Springer, \bibinfo{pages}{740--755}.
\newblock


\bibitem[Lundberg and Lee(2017)]%
        {lundberg2017unified}
\bibfield{author}{\bibinfo{person}{Scott~M Lundberg} {and} \bibinfo{person}{Su-In Lee}.} \bibinfo{year}{2017}\natexlab{}.
\newblock \showarticletitle{A unified approach to interpreting model predictions}.
\newblock \bibinfo{journal}{\emph{Advances in neural information processing systems}}  \bibinfo{volume}{30} (\bibinfo{year}{2017}).
\newblock


\bibitem[Ribeiro et~al\mbox{.}(2016)]%
        {ribeiro2016should}
\bibfield{author}{\bibinfo{person}{Marco~Tulio Ribeiro}, \bibinfo{person}{Sameer Singh}, {and} \bibinfo{person}{Carlos Guestrin}.} \bibinfo{year}{2016}\natexlab{}.
\newblock \showarticletitle{" Why should i trust you?" Explaining the predictions of any classifier}. In \bibinfo{booktitle}{\emph{Proceedings of the 22nd ACM SIGKDD international conference on knowledge discovery and data mining}}. \bibinfo{pages}{1135--1144}.
\newblock


\bibitem[Selvaraju et~al\mbox{.}(2017)]%
        {selvaraju2017grad}
\bibfield{author}{\bibinfo{person}{Ramprasaath~R Selvaraju}, \bibinfo{person}{Michael Cogswell}, \bibinfo{person}{Abhishek Das}, \bibinfo{person}{Ramakrishna Vedantam}, \bibinfo{person}{Devi Parikh}, {and} \bibinfo{person}{Dhruv Batra}.} \bibinfo{year}{2017}\natexlab{}.
\newblock \showarticletitle{Grad-cam: Visual explanations from deep networks via gradient-based localization}. In \bibinfo{booktitle}{\emph{Proceedings of the IEEE international conference on computer vision}}. \bibinfo{pages}{618--626}.
\newblock


\bibitem[Shrikumar et~al\mbox{.}(2017)]%
        {pmlr-v70-shrikumar17a}
\bibfield{author}{\bibinfo{person}{Avanti Shrikumar}, \bibinfo{person}{Peyton Greenside}, {and} \bibinfo{person}{Anshul Kundaje}.} \bibinfo{year}{2017}\natexlab{}.
\newblock \showarticletitle{Learning Important Features Through Propagating Activation Differences}. In \bibinfo{booktitle}{\emph{Proceedings of the 34th International Conference on Machine Learning}} \emph{(\bibinfo{series}{Proceedings of Machine Learning Research}, Vol.~\bibinfo{volume}{70})}, \bibfield{editor}{\bibinfo{person}{Doina Precup} {and} \bibinfo{person}{Yee~Whye Teh}} (Eds.). \bibinfo{publisher}{PMLR}, \bibinfo{pages}{3145--3153}.
\newblock
\urldef\tempurl%
\url{https://proceedings.mlr.press/v70/shrikumar17a.html}
\showURL{%
\tempurl}


\bibitem[Simonyan et~al\mbox{.}(2014)]%
        {simonyan2014deepinsideconvolutionalnetworks}
\bibfield{author}{\bibinfo{person}{Karen Simonyan}, \bibinfo{person}{Andrea Vedaldi}, {and} \bibinfo{person}{Andrew Zisserman}.} \bibinfo{year}{2014}\natexlab{}.
\newblock \bibinfo{title}{Deep Inside Convolutional Networks: Visualising Image Classification Models and Saliency Maps}.
\newblock
\newblock
\showeprint[arxiv]{1312.6034}~[cs.CV]
\urldef\tempurl%
\url{https://arxiv.org/abs/1312.6034}
\showURL{%
\tempurl}


\bibitem[Simonyan and Zisserman(2014)]%
        {simonyan2014very}
\bibfield{author}{\bibinfo{person}{Karen Simonyan} {and} \bibinfo{person}{Andrew Zisserman}.} \bibinfo{year}{2014}\natexlab{}.
\newblock \showarticletitle{Very deep convolutional networks for large-scale image recognition}.
\newblock \bibinfo{journal}{\emph{arXiv preprint arXiv:1409.1556}} (\bibinfo{year}{2014}).
\newblock


\bibitem[Springenberg et~al\mbox{.}(2014)]%
        {springenberg2014striving}
\bibfield{author}{\bibinfo{person}{Jost~Tobias Springenberg}, \bibinfo{person}{Alexey Dosovitskiy}, \bibinfo{person}{Thomas Brox}, {and} \bibinfo{person}{Martin Riedmiller}.} \bibinfo{year}{2014}\natexlab{}.
\newblock \showarticletitle{Striving for simplicity: The all convolutional net}.
\newblock \bibinfo{journal}{\emph{arXiv preprint arXiv:1412.6806}} (\bibinfo{year}{2014}).
\newblock


\bibitem[Sundararajan et~al\mbox{.}(2017)]%
        {sundararajan2017axiomatic}
\bibfield{author}{\bibinfo{person}{Mukund Sundararajan}, \bibinfo{person}{Ankur Taly}, {and} \bibinfo{person}{Qiqi Yan}.} \bibinfo{year}{2017}\natexlab{}.
\newblock \showarticletitle{Axiomatic attribution for deep networks}. In \bibinfo{booktitle}{\emph{International conference on machine learning}}. PMLR, \bibinfo{pages}{3319--3328}.
\newblock


\bibitem[Szegedy et~al\mbox{.}(2016)]%
        {szegedy2016rethinking}
\bibfield{author}{\bibinfo{person}{Christian Szegedy}, \bibinfo{person}{Vincent Vanhoucke}, \bibinfo{person}{Sergey Ioffe}, \bibinfo{person}{Jon Shlens}, {and} \bibinfo{person}{Zbigniew Wojna}.} \bibinfo{year}{2016}\natexlab{}.
\newblock \showarticletitle{Rethinking the inception architecture for computer vision}. In \bibinfo{booktitle}{\emph{Proceedings of the IEEE conference on computer vision and pattern recognition}}. \bibinfo{pages}{2818--2826}.
\newblock


\bibitem[Zeiler(2014)]%
        {zeiler2014visualizing}
\bibfield{author}{\bibinfo{person}{MD Zeiler}.} \bibinfo{year}{2014}\natexlab{}.
\newblock \showarticletitle{Visualizing and Understanding Convolutional Networks}. In \bibinfo{booktitle}{\emph{European conference on computer vision/arXiv}}, Vol.~\bibinfo{volume}{1311}.
\newblock


\end{thebibliography}
}

\newpage
\appendix

\section{Appendix}

\subsection {Illustration of rank-based redaction for ResNet-50 and Inception-v3}

\begin{figure}[!h]
    \centering
        \includegraphics[scale=0.68]{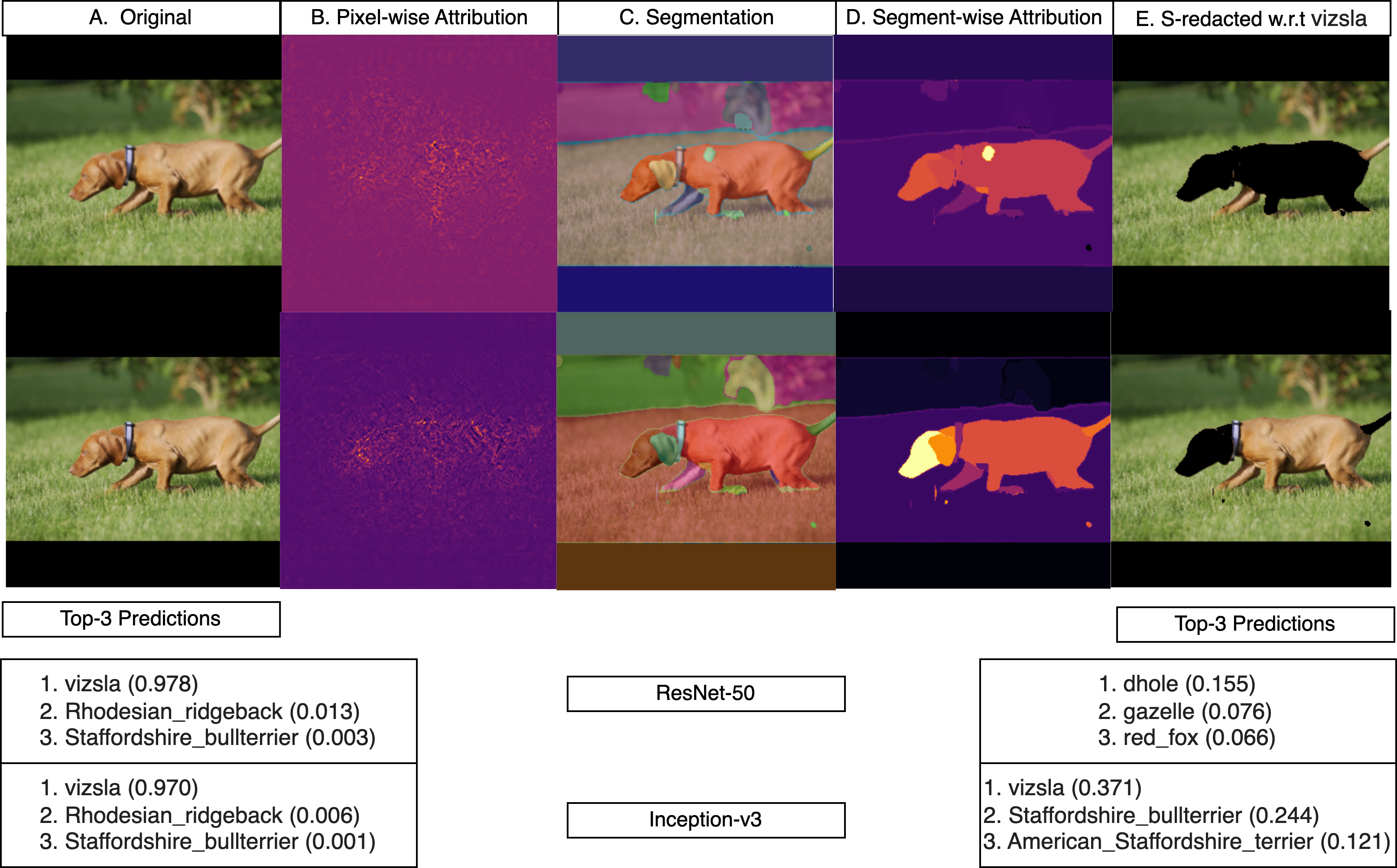}
        \caption{Illustration of rank-based redaction for different models ({\em Top row : ResNet-50 , Bottom row : Inception-v3}) using the same image and following a similar pipeline as Figure \ref{fig1}.}
    \label{overview_resnet_inception}
\end{figure}

Figure \ref{overview_resnet_inception} illustrates rank-based redactions for ResNet-50 and Inception-v3 for a sample image. Observe that an image can have differing  $S$-redactions for different models.  


\subsection {Two additional algorithms for finding $\delta$-disjoint attributions}
\label{othertype1}

\subsubsection{Algorithm 2} 
\label{type1algo2}

As we perform the steps mentioned in Section \ref{Redaction}, for labels $l_1$ and $l_2$, we obtain two identical sets of segments that are ranked within their set based on segment-wise attribution method \cite{kapishnikov2019xrai}. Now that we have two sets of segments, we pick each segment from label $l_1$'s set based on their rank and redact by sequentially accumulating them to form an $S_1$-redaction until the prediction of corresponding class label $l_1$ goes down to at most {\em $\delta$}$p_1$ while the $l_2$ prediction stays above (1-{\em $\delta$})$p_2$ where $p_1$ and $p_2$ are the softmax probability of original image of labels $l_1$ and $l_2$ respectively. This step is repeated on $l_2$'s set to obtain $S_2$ redaction. Redacting based on their rank in corresponding sets allows us to get the $S_1$  and $S_2$ redactions with small number of segments. These two redactions might not satisfy Definition \ref{type1def} as they may not be disjoint. We then discard the intersection segments or reassign them one-by-one to either $S_1$ or $S_2$ redactions based on their importance to the respective labels. This step makes the $S_1$  and $S_2$ redactions disjoint and satisfies Definition \ref{type1def}. 


\subsubsection{Algorithm 3 (which also serves as a heuristic verifier for Definition \ref{type2def})}
\label{type1algo3}

Given a set of segments $E$ that are obtained after segmentation, we would like to curate two disjoint sets of segments that are $\delta$ attributions for each label and can generate $S$-redactions that satisfy Definition \ref{type1def}. To generate $\delta$ attribution set of segments for label $l_1$,  we start with an empty set $A$ and execute the following algorithm.
\begin{enumerate}

\item Select the most important segment $s_i$ and pop it out of $E$ and push it into $A$.
\item Generate $D$ with the set of segments that are adjacent to any segment in $A$. 
\item Pop out next most important segment from $D$ and push it into $A$. 
\item Repeat the process from $Step2$ until we end up with no important segment in $D$. 
\item Repeat from $Step1$ and start with a different segment until we are left with no important segments for label $l_1$. This process is repeated to generate $\delta$ attribution set of segments for label $l_2$.
    
\end{enumerate}

How do we choose most important segment? To calculate the importance score of segment $s_i$ we redact the segment $s_i$ from $A$-redacted image(image with all segments from $A$ redacted) and check the percentage drop in $l_1$ and $l_2$ prediction values from $A$-redaction to {$A\cup \{s_i$\}} redaction. The difference between $l_1$ percentage drop and $l_2$ percentage drop is the importance score for the segment $s_i$ during that step.

We end up with $S_1$ and $S_2$ redactions that are not disjoint, and discarding the intersection gives us two disjoint redactions. These redactions are then verified to check if they satisfy Definition \ref{type1def} and corresponding $\delta$ attributions are used as certificates to validate the image later.

The limitation of this algorithm is that , unlike attribution based algorithms from Sections \ref{type1} \& \ref{type1algo2}, it does not provide the redactions with a small number of segments, since it does not pick the segments based on the ranks generated by the attribution algorithms.

\begin{figure}[!h]
    \centering
        \includegraphics[scale=0.7]{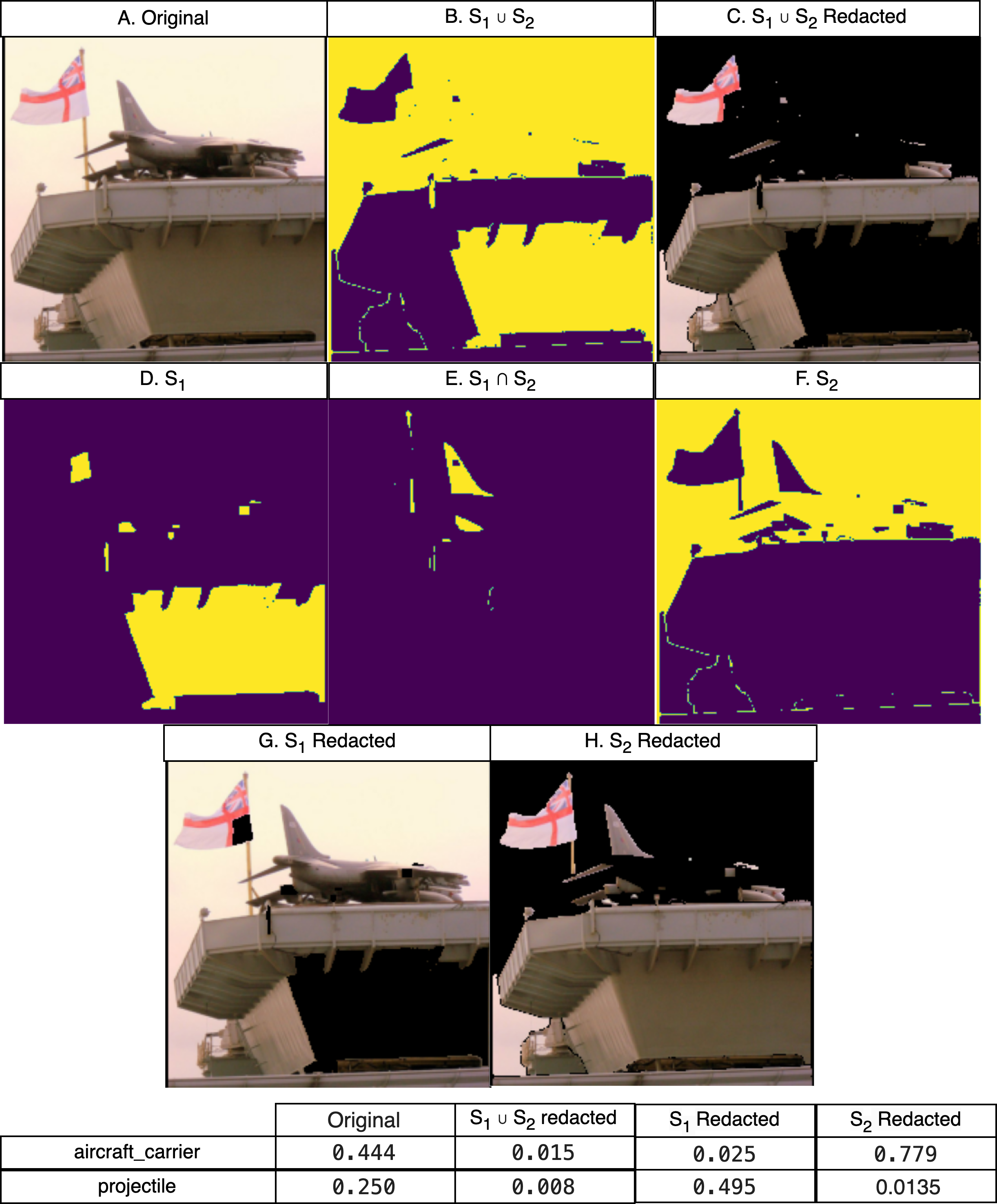}
        \caption{ Example illustrating Algorithm 3 as a verifier for Definition \ref{type2def}. Here we take a sample image which satisfies the definition of $\delta$-disjoint attributions for a pair of classes. We then suppose $S=S_1 \cup S_2$ and challenge the verifier by offering $S$ as a certificate for $\delta$-overlapping attributions. We demonstrate that the verifier indeed flags $S$ as an incorrect certificate for $\delta$-overlapping attributions.  {\bf A. }An image  with labels {\tt aircraft\_carrier} and {\tt projectile} which satisfies Definition \ref{type1def} on ResNet-50 model. {\bf B\&C. } $S=S_1 \cup S_2$ acting as a purported  certificate for Definition \ref{type2def}, with $\delta=0.2$, for both {\tt aircraft\_carrier} and {\tt projectile} classes.  {\bf D, E \& F.} Algorithm \ref{type1algo3} breaks the set $S$ to become $S_1, S_1 \cap S_2, S_2$ in {\em D, E \& F} respectively. {\bf G \& H. }$S_1$ and $S_2$ redaction images formed using {\em D \& F} respectively. {\em G \& H} together satisfies Definition \ref{type1def} with $\delta=0.2$ and hence the verifier rejects the certificate $S$ for Definition \ref{type2def}.}
    \label{verification}
\end{figure}

This algorithm also acts as a heuristic verifier for Definition \ref{type2def}. Provided with set of segments $S=S_1 \cup S_2$ as $\delta$ overlapping attribution certificate from Section \ref{type2}, if the image and labels correspond to $\delta$ disjoint from Section \ref{type1}, this algorithm segregates the segments into $S_1$ and $S_2$ sets that correspond to each labels satisfying Definition \ref{type1def}. These $S_1$ and $S_2$ sets invalidate the $\delta$ attribution certificate. Usage of this algorithm can remove the overhead of attribution. User performing the verification need not have the knowledge of algorithm that generated the certificate. An example is demonstrated in Figure \ref{verification}.

\clearpage

\subsubsection{Additional examples on VGG-16, ResNet-50 and Inception-v3}
\label{additional_examples}

\begin{figure}[!ht]
    \centering
        \includegraphics[scale=0.45]{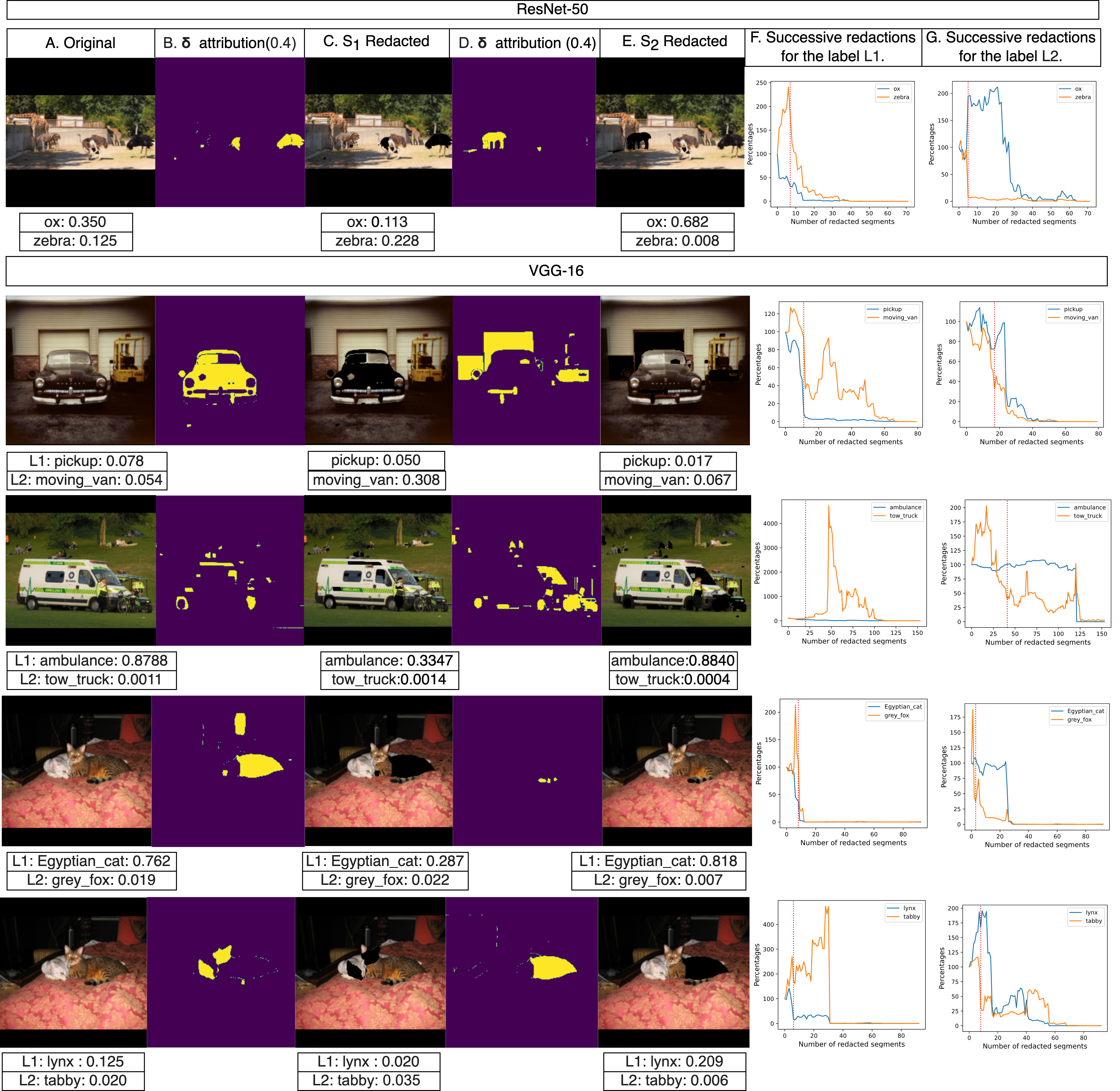}
        \caption{Illustrations of $\delta$-disjoint attributions with $\delta$=0.4 on ResNet-50 and VGG-16 using various images following a similar pipeline as Figure \ref{fig2}. Image used in the first row is not from the ImageNet validation dataset.}
    \label{type1_other}
\end{figure}

\begin{figure}[!ht]
    \centering
        \includegraphics[scale=0.70]{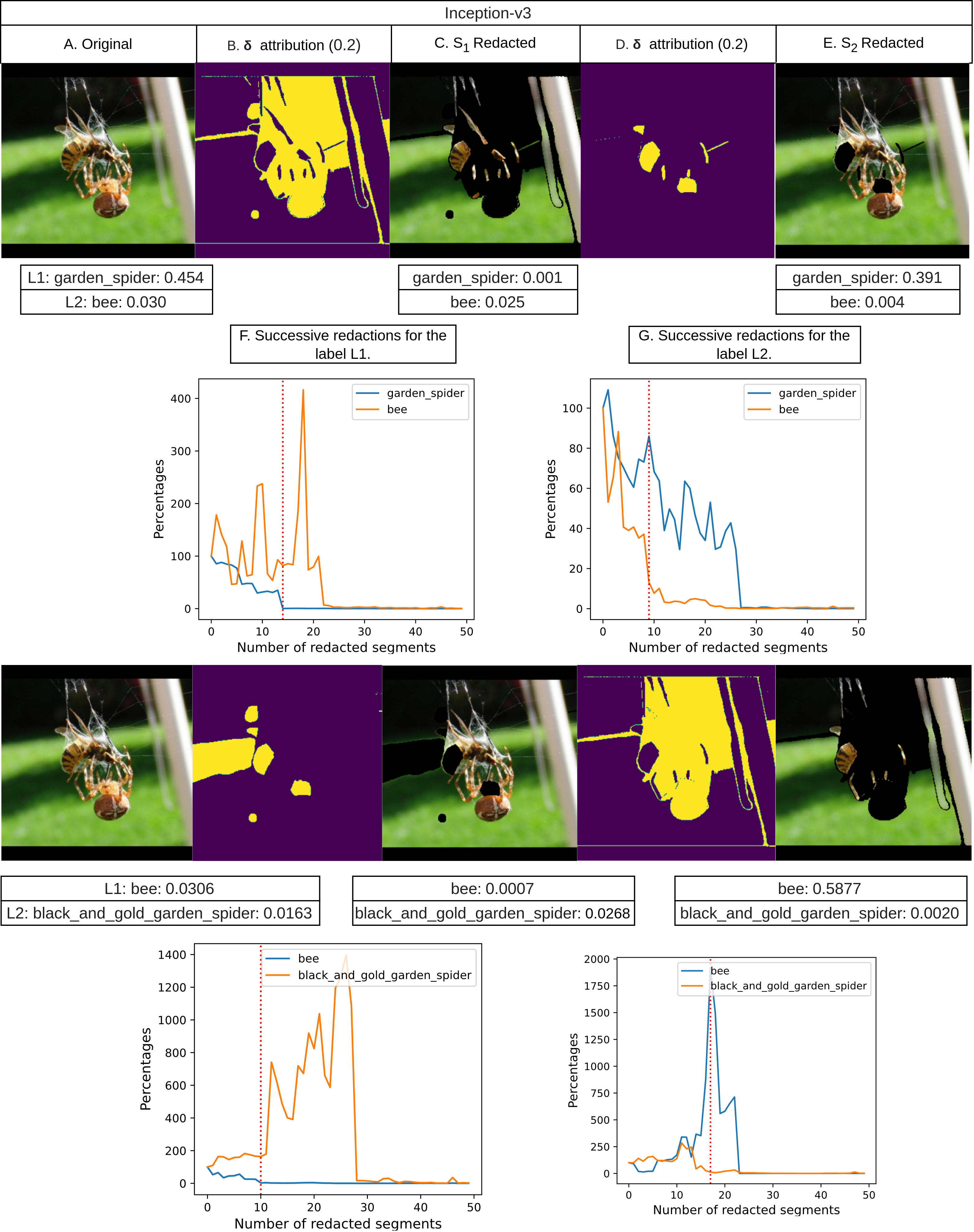}
        \caption{Illustrations of $\delta$-disjoint attributions with $\delta$=0.4 on Inception-v3 for two pairs of labels using a single image following a similar pipeline as Figure \ref{fig2}. Observe that the image indeed has both a bee and a spider. The image is from the {\tt garden\_spider} class from the ImageNet validation dataset.}
    \label{type1_spider} 
\end{figure}

\begin{figure}[!ht]
    \centering
        \includegraphics[scale=0.60]{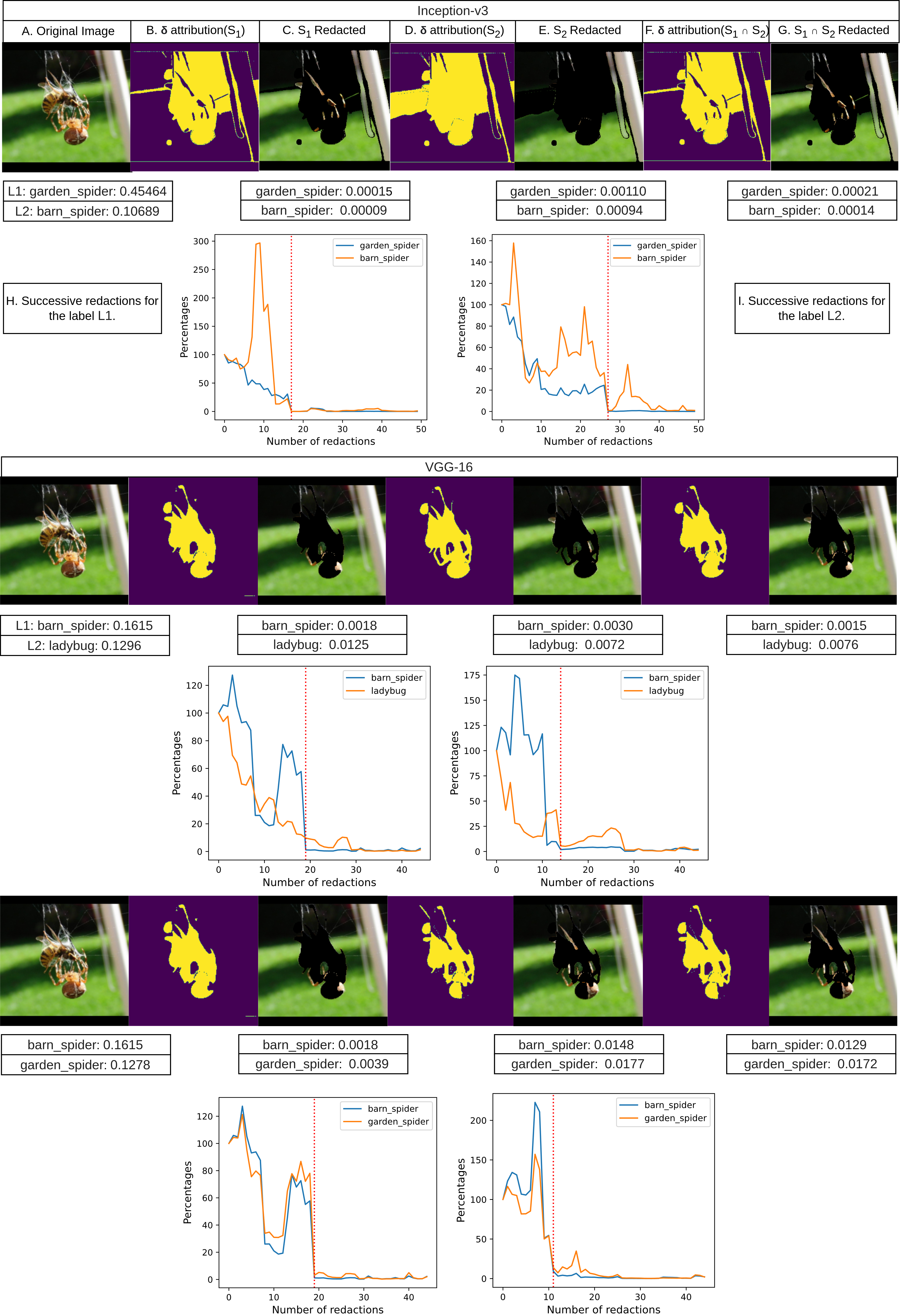}
        \caption{Illustrations of $\delta$-overlapping attributions with $\delta$=0.2 on Inception-v3 and VGG-16 for various labels using the same image as Figure \ref{type1_spider} and following a similar pipeline as Figure \ref{fig3}. }
    \label{type2_spider}
\end{figure}

Illustrations of $\delta$-disjoint and $\delta$-overlapping 
attributions and their corresponding redactions performed on various images are shown in Figures \ref{type1_other}, \ref{type1_spider}, and \ref{type2_spider}. Note that the label pairs chosen in these figures are picked from top-5 rather than top-2.

We mention an example in Figure \ref{type1_other} (VGG-16, example 1), wherein the method flags it as a $\delta$-disjoint attribution, even though a human inspection shows that no object corresponding to the class {\tt moving\_van} is present. This is a limitation, as previously mentioned, even though the method shows that differing segments cause the $\delta$ attributions for the two classes.

Figure \ref{type1_spider} and \ref{type2_spider} demonstrate our method on a challenging example, which contains both a spider and a bee. In a pair of classes that correspond to a spider and a bee, it is classified as a $\delta$-disjoint attribution; however with two spider class labels, it is classified as a $\delta$-overlapping attribution.  Specifically, in Figure \ref{type1_spider}, it is observed that for Inception-v3, the image is categorized into $\delta$-disjoint attribution for pair of labels ({\tt garden\_spider}, {\tt bee}) and ({\tt bee}, {\tt black\_and\_gold\_garden\_spider}), whereas the same image for labels ({\tt garden\_spider}, {\tt barn\_spider}) is categorized into $\delta$-overlapping attribution as shown in Figure \ref{type2_spider}.

\subsection{Running time estimates for our methods}
\label{runtime}
All our experiments were run on an Apple Macbook Pro with M1 Pro chip, 16GB RAM, running macOS 12.1, and the running time estimates below correspond to this hardware.

\noindent
SAM segmentation algorithm: 127 seconds per image.\\
Pixel-wise attribution and Segment-wise attribution: 64 seconds for a pair of chosen labels.\\
Algorithm mentioned in Section \ref{type1}: 15 seconds for a pair of chosen labels.\\
Algorithm mentioned in Section \ref{type2}: 16 seconds for a pair of chosen labels.\\
Algorithm mentioned in Section \ref{type1algo2}: 16 seconds for a pair of chosen labels.\\
Algorithm mentioned in Section \ref{type1algo3}: 900 seconds for a pair of chosen labels.

\newpage

\end{document}